\documentclass[preprint,12pt]{elsarticle}

\usepackage{amssymb}
\usepackage{hyperref}
\usepackage{graphicx}
\usepackage{subcaption}
\usepackage{listings}

\usepackage{siunitx}
\journal{Science of Computer Programming}

\begin{document}

\begin{frontmatter}



\title{AmbieGen: A Search-based Framework for Autonomous Systems Testing}


\author{Dmytro Humeniuk}
\author {Foutse Khomh}
\author {Giuliano Antoniol}
\address{Polytechnique Montréal, 2500 Chemin de Polytechnique,  QC H3T 1J4, Montréal, Canada }
\begin{abstract}
Thorough testing of safety-critical autonomous systems, such as self-driving cars, autonomous robots, and drones, is essential for detecting potential failures before deployment. One crucial testing stage is model-in-the-loop testing, where the system model is evaluated by executing various scenarios in a simulator. However, the search space of possible parameters defining these test scenarios is vast, and simulating all combinations is computationally infeasible.
To address this challenge, we introduce AmbieGen, a search-based test case generation framework for autonomous systems. AmbieGen uses evolutionary search to identify the most critical scenarios for a given system, and has a modular architecture that allows for the addition of new systems under test, algorithms, and search operators. Currently, AmbieGen supports test case generation for autonomous robots and autonomous car lane keeping assist systems. In this paper, we provide a high-level overview of the framework's architecture and demonstrate its practical use cases.

\end{abstract}

\begin{keyword}
evolutionary search \sep autonomous systems \sep self driving cars \sep autonomous robots \sep neural network testing
\end{keyword}

\end{frontmatter}


\section*{Metadata}
\label{}
The project metadata is presented in Table \ref{tab:metadeta}.

\begin{table}[h!]
\begin{tabular}{|l|p{6.5cm}|p{6.5cm}|}
\hline
\textbf{Nr.} & \textbf{Code metadata description} & \textbf{Please fill in this column} \\
\hline
C1 & Current code version & v0.1.0 \\
\hline
C2 & Permanent link to code/repository used for this code version & For example: \url{https://github.com/swat-lab-optimization/AmbieGen-tool} \\
\hline
C3  & Permanent link to Reproducible Capsule & \url{https://codeocean.com/capsule/1741442/tree} \\ 
\hline
C4 & Legal Code License   & MIT license (MIT) \\ 
\hline
C5 & Code versioning system used & git \\
\hline
C6 & Software code languages, tools, and services used & python \\
\hline
C7 & Compilation requirements, operating environments and dependencies & indicated in requirements.txt\\
\hline
C8 & If available, link to developer documentation/manual & \url{https://github.com/swat-lab-optimization/AmbieGen-tool/blob/master/README.md} \\
\hline
C9 & Support email for questions & dmytro.humeniuk@polymtl.ca \\
\hline
\end{tabular}
\caption{Code metadata (mandatory)}
\label{tab:metadeta}
\end{table}

\section{Motivation and significance}

Autonomous systems, including autonomous vehicles, robots, or drones can provide a number of benefits such as driving assistance, high-risk zone exploration, and aid in rescue operations. At the same time, these are safety-critical systems and it is very important to ensure they are robust to unseen environments and conditions. This can be done by thorough testing prior to their deployment. Typically, at the initial development stages model-in-the-loop testing is performed \cite{bruggner2021model}, where the system is tested in a simulation environment.
Given the complexity of autonomous systems, the number of potential test scenarios is vast and exhaustive execution is not feasible.
For example, an autonomous vehicle scenario could involve a variety of parameters such as road topology, the movement and behavior of other vehicles and pedestrians, traffic signs, weather conditions, etc. We surmise that in order to identify the most critical scenarios for a given system, application of search algorithms is necessary.

In this work, we propose AmbieGen, a search based framework for generating adversarial test scenarios for autonomous systems. By leveraging evolutionary search AmbieGen allows to find challenging and diverse test scenarios.

The problem of identifying critical scenarios for a system has been addressed in several previous works on falsifying temporal logic requirements of cyber-physical systems, such as S-Taliro \cite{annpureddy2011s}, Breach \cite{donze2010breach}, and ARIsTEO \cite{menghi2020approximation}. These works typically consider falsifying a model of the system that takes a set of input signals and produces a set of output signals.

In our work, we focus on testing autonomous systems for which the input signals are complex and may include data from various sensors and cameras. Generating a valid combination of falsifying input signals (such as lidar readings and RGB camera readings) directly would be challenging. Therefore, we propose a method for generating test cases that specify a virtual environment for the autonomous system, rather than the input signals. The input signals are generated in the virtual environment during simulation based on the actions of the autonomous agent.

Several approaches have been proposed for generating virtual environments for testing autonomous driving and robotics systems, including AsFault \cite{gambi2019automatically}, Frenetic \cite{castellano2021frenetic}, DeepJanus \cite{riccio2020model}, DeepHyperion \cite{zohdinasab2021deephyperion} and others presented at the SBST 2021 \cite{panichella2021sbst} and SBST 2022 \cite{gambi2022sbst} tool competitions.

The tool we present in this paper, AmbieGen, is the winner of SBST 2022 tool competition. It could produce the biggest number of diverse fault revealing scenarios for an autonomous vehicle lane keeping assist system (LKAS) given a limited time budget. More details about the search algorithm implementation can be found in our research paper \cite{HUMENIUK2022106936}. In our work we have shown that the simplified model of the system can be effective in guiding the search for producing the test scenarios for the full, simulator based, model of the system. 

Our framework can be used for further research in the search algorithms, search operator and fitness function design for autonomous systems adversarial testing. We built the framework to be modular, and thus easily customizable. By referring to project documentation as well as the example implementations we provide, researchers can specify their own test scenario generation problems, fitness functions, crossover and mutation operators. This tool is developed in Python and can be easily run as a python package. More instructions and examples are provided in the \href{https://github.com/swat-lab-optimization/AmbieGen-tool}{AmbieGen repository}.




\section{Software description}
In this work, we present AmbieGen, an open-source Python framework that utilizes evolutionary search for the generation of test scenarios for autonomous systems. Currently, AmbieGen supports the creation of test scenarios for lane keeping assist systems (LKAS) in autonomous vehicles and for autonomous robots navigating a closed room with obstacles.

\begin{figure}[h]
\centering
\includegraphics[scale=0.55]{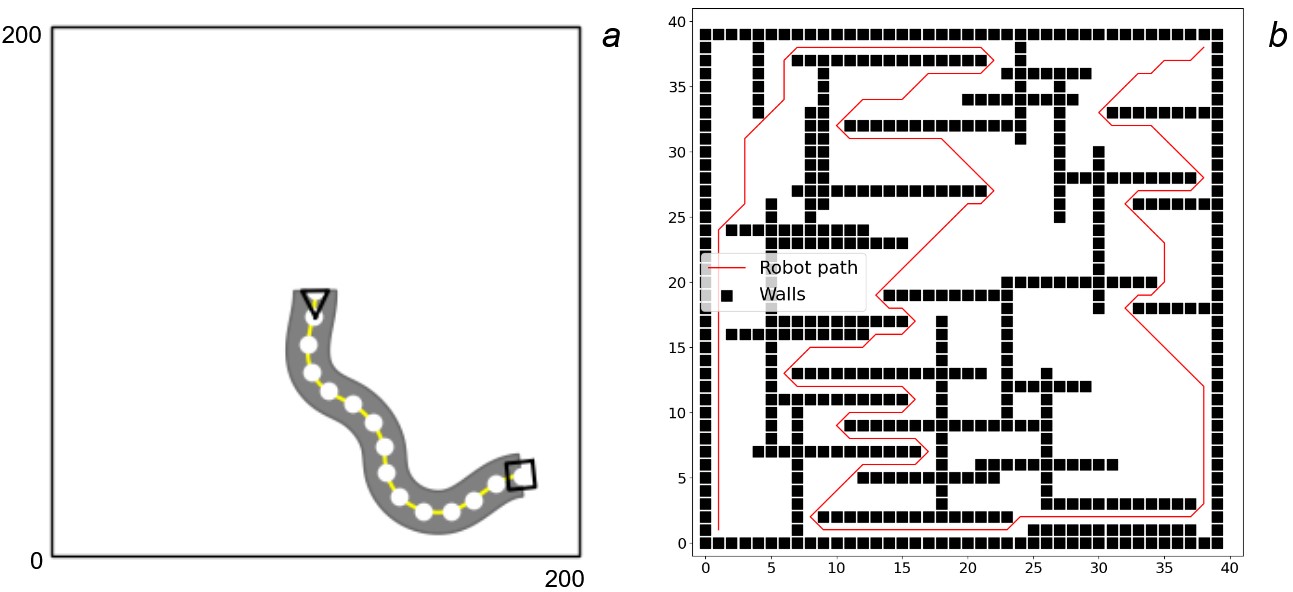}

\caption{An example of the test case for LKAS system (a) and an autonomous robot (b). The x-axis represents the map length in meters, and the y-axis represents the map width in meters.}
\label{fig:examples}
\end{figure}

The test scenarios for LKAS in vehicles are designed to challenge the system with various road topologies, while the scenarios for autonomous robots involve navigating a closed room with obstacles. Examples of the generated scenarios can be seen in Figure \ref{fig:examples}.

\subsection{Software architecture}
\begin{figure}[h]
\centering
\includegraphics[scale=0.58]{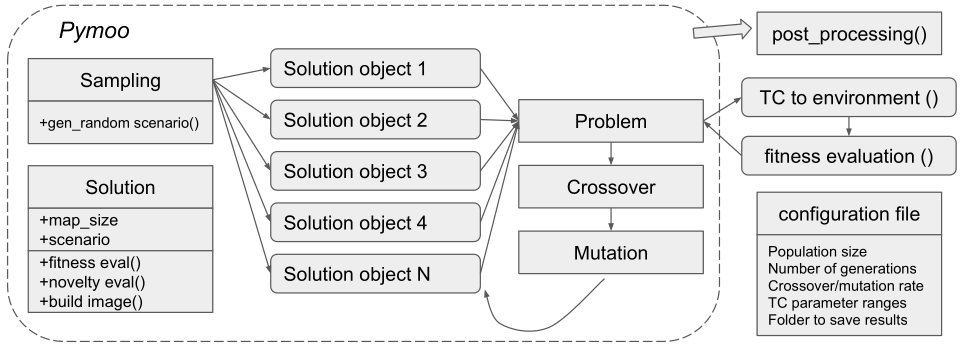}
\caption{AmbieGen architecture}
\label{fig:architechture}
\end{figure}
This subsection provides a detailed description of the software implementation of AmbieGen. The key components of AmbieGen are illustrated in Figure \ref{fig:architechture}, which are common components for implementing evolutionary search. We use the Pymoo framework \cite{pymoo} to implement the search algorithms. The most important modules and classes are outlined below: 

\begin{itemize}
    
    \item \textit{Solution} - this is one of the most important classes, which contains all the necessary attributes and functions needed to represent the candidate solution of the algorithm. It should contain a \textit{scenario} attribute with the list of test case parameters, function for fitness evaluation, novelty calculation, as well as, optionally, image generation. 
    \item \textit{Sampling} - this is the class for initial population generation. At the output it  provides $N$ instances of the Solution class, with the initialized \textit{scenario} attribute, defining the test scenario. Typically the test scenario is represented by a two dimensional array, randomly initialized based on the minimum and maximum values of the test case parameters, defined in the configuration file. Each column of the array corresponds to some part of the environment. More information about the representation of the test scenarios that we used can be found in the repository page as well as in our research article.
    \item \textit{Problem} - in this class, we define the logic for evaluating the fitness of each solution. For single-objective search (using GA), we specify the fitness function for evaluating the scenario fault revealing power. For two-objective search (using NSGA-II), we define two objectives: fault revealing power and novelty calculation. The novelty objective is calculated as the average novelty of a given test scenario relative to the 5 solutions with the highest fault revealing power fitness. If the problem has any constraints, such as a minimum required fitness value, they should also be specified in this class.
    \item \textit{TC to environment} - this is a function to transform the test case (TC) encoded as a 2D array of parameters, to the input format suitable for the system model. For example, for the LKAS problem, the model input is a list of the 2D coordinates of points, defining the road topology. The test case itself is represented as a sequence of transformations needed to perform to obtain the points. For the autonomous robot the test scenario is represented as a sequence of parameters describing the 2D map with obstacles. The \textit{TC to environment} module is used to create a 2D bitmap from the given parameters. The bitmap is given as the input to the autonomous robot model, which runs a planning algorithm to find the shortest path between the start and goal location.
    \item \textit{fitness evaluation} - a function to calculate the fitness i.e fault revealing power of the scenario. It takes the output of the \textit{TC to environment} function as the input and execute the system model. It collects the data about the model behaviour during execution and computes the fitness score. For the LKAS system, the fitness is defined by the biggest deviation from the lane center and for the autonomous robot - by the length of the path to reach the goal.
    \item \textit{Crossover} - in this class the crossover operator is defined. Currently we are using a one point crossover, which can be applied to fixed and variable length solutions. 
    \item \textit{Mutation} - in this class the mutation operator is implemented. We have 2 types of mutations: exchange and change of variable. In exchange mutation, two randomly selected columns of the test case are exchanged. In the case of the road topology, it would correspond to exchanging the positions of two random road segments. In change of variable mutation,  a randomly selected parameter value in the test case matrix is changed. In the road topology example it could correspond to the change of the length of one of the straight road segments.
    \item \textit{post processing} - The post-processing module of our framework includes several functions for handling the test suite and its metadata. The function $get\_test\_suite()$ retrieves the test suite, $get\_stats()$ retrieves metadata such as fitness and novelty scores, and $save\_tcs\_images()$ saves the images of the test cases. The size of the test suite, denoted as $T$, can be specified in the configuration file. In our experiments, $T$ was typically set to 30, representing the best solutions found by the algorithm.

Metadata for the test suite includes the fitness of the top $T$ solutions, their novelty (calculated as the average novelty between all pairs of scenarios in the test suite), and the convergence (best solution fitness found at each epoch). The post-processing module also includes a $compare.py$ script for comparing the results of different algorithms, using the collected metadata to generate convergence plots and fitness and diversity boxplots.
    \item \textit{configuration file} - finally we have a configuration file, where the parameters of the algorithm, such as: the population size, the number of generations, crossover/mutation rate, and the test suite size are defined. Users should also specify the allowable ranges for  the test case parameters and the paths for saving the resulting test suite and its metadata.

\end{itemize}

Currently, when adding a new problem, one should provide the implementation of each of the modules as well as the \textit{TC to environment} and \textit{fitness evaluation} functions. We are working on reducing the number of additional implementations needed. Our framework includes the implementation of all the modules for the LKAS and autonomous robot test case generation problems.

\subsection{Software functionalities}

AmbieGen public version 0.1.0 as presented in this paper offers the following major functionalities:
\begin{itemize}
    \item \textit{Autonomous vehicle LKAS system testing}: generating scenarios, represented as a list of 2D coordinates defining the road topology.
    \item \textit{Autonomous robot testing}: generating scenarios, represented as the 2D bitmap, defining obstacle locations in a fixed sized map.

    \item \textit{Search-based generation}: our framework provides options for search-based test suite generation, including random search, single-objective genetic algorithm (GA), and two-objective genetic algorithm (NSGA-II). The search algorithms are implemented using the Pymoo framework \cite{pymoo}, and can be easily extended to support additional algorithms supported by Pymoo with minor modifications.

The single-objective GA optimizes the test suite for scenario fault revealing power, while the two-objective NSGA-II optimizes for both fault revealing power and diversity. As demonstrated in our previous work \cite{HUMENIUK2022106936}, the two-objective algorithm allows to produce a more diverse set of test scenarios compared to the single-objective search. 
    \item \textit{Experiment data tracking}: AmbieGen tracks the results of each experiment and saves them in a user-defined location. The saved data includes the $T$ (as determined by the user) best test scenarios identified based on their fitness or crowding distance, as well as their associated metadata such as fitness, average diversity, and visualizations. This allows for easy analysis and comparison of the results of different experiments. 
\end{itemize}

\subsection{Use cases of the software}
\begin{figure}[h!]
\centering

\includegraphics[scale=0.65]{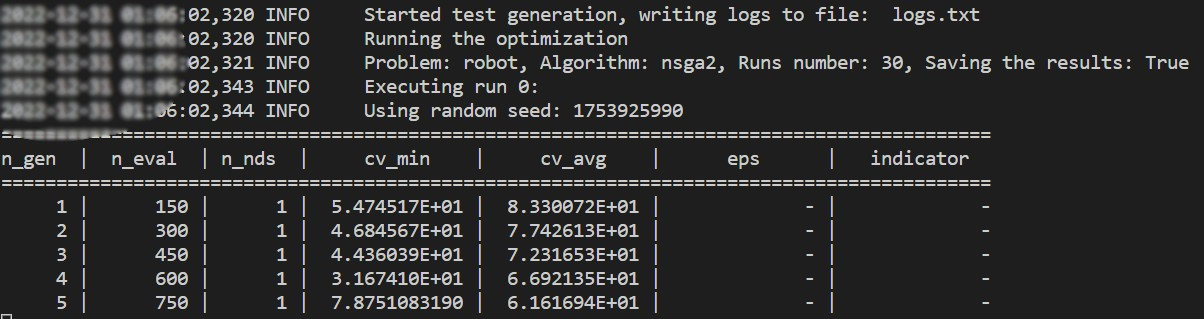}
\caption{Printouts during the search}
\label{fig:config}
\end{figure}

In this subsection we provide an illustrative example of how to use AmbieGen to generate test cases for an autonomous robot planning algorithm testing.
Suppose we want to perform 30 runs of the NSGA-II algorithm with 150 individuals and 200 generations to evaluate this configuration. We want to save the generated test cases, their illustrations as well as their metadata, such as fitness and diversity. 
Below you can see the configuration file entries with the parameters we chose for the genetic algorithm and well as the path to save the experiment results:

\begin{lstlisting}
ga = {"pop_size": 150, "n_gen": 200, "mut_rate": 0.4, "cross_rate": 0.9,
"test_suite_size": 30 }
files = {"stats_path": "stats", "tcs_path": "tcs", "images_path": images"}
\end{lstlisting}

Now we are ready to start the test case generation. We can launch AmbieGen with the following command and parameters:
\begin{lstlisting}
python optimize.py --problem="robot" --algo="nsga2" --runs=30 \\
--save_results=True
\end{lstlisting}
The search will start and you could see some printouts, such as in Fig. \ref{fig:config} with the current number of generation (n\_gen), number of evaluations (n\_eval), constraint violation (cv\_min), number of non-dominant solution for NSGA-II algorithm (n\_nds) and the best solution found (f\_opt) for GA algorithm. More details about the printed information can be found on the Pymoo page (\url{https://pymoo.org/interface/display.html}).

\begin{figure}[h!]
\centering
\includegraphics[scale=0.65]{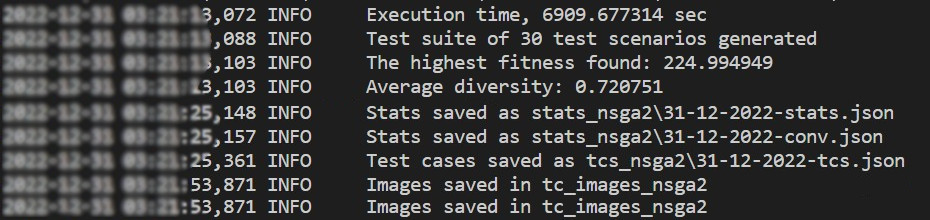}
\caption{Successful run confirmation}
\label{fig:results}
\end{figure}

After a successful run, you will see the confirmation about the run execution time,  saved test cases, their metadata and the images, as in Fig. \ref{fig:results}

\begin{figure}[h!]
\begin{subfigure}{.5\textwidth}
  \centering
  \includegraphics[scale=0.98]{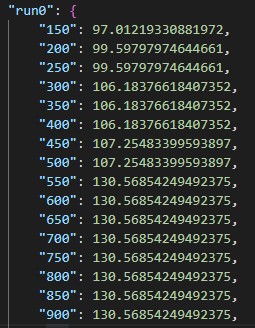}  
  \caption{Scenario fitness convergence}
  \label{fig:conv}
\end{subfigure}
\begin{subfigure}{.5\textwidth}
  \centering
  \includegraphics[scale=0.94]{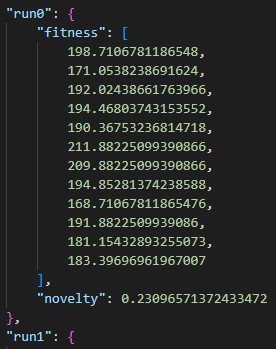}  
  \caption{Final test suite fitness and diversity}
  \label{fig:fit}
\end{subfigure}
\caption{Metadata for the generated scenarios}
\label{fig:results2}
\end{figure}

\begin{figure}[h!]
\centering
\includegraphics[scale=0.65]{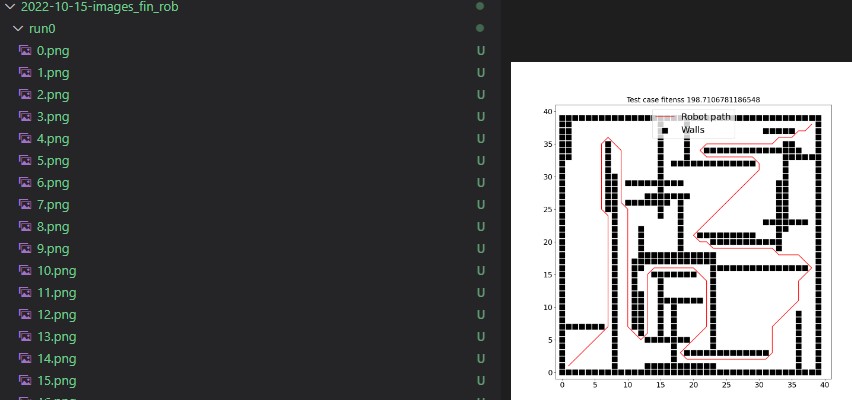}
\caption{Images of the generated scenarios}
\label{fig:images}
\end{figure}

 In Fig. \ref{fig:results2} you can see examples of the metadata saved, such as the algorithm convergence \ref{fig:conv} (the best fitness value at each generation in the format "evaluation number": best fitness found), the fitness of the test cases in the test suite as well as their average diversity i.e., novelty \ref{fig:fit}. Novelty is calculated as the average diversity of all of the pairs of the test cases in the test suite.
In Fig. \ref{fig:images} we show an example of the test case images saved for a particular run.

Finally, let us suppose we also want to run a random search with the same evaluation budget to be able to compare the performance of our configuration of NSGA-II algorithm to some baseline. We can run the random search by executing the following command:
\begin{lstlisting}
python optimize.py --problem="robot" --algo="random" --runs=30 \\
--save_results=True
\end{lstlisting}
The random search will be run and the metadata saved, as in the previous case.
Now we can compare the results produced by the two different search algorithms via executing the following command:
\begin{lstlisting}
python compare.py --stats_path="stats_nsga2" "stats_random" \\
--stats_names "NSGA-II" "Random"
\end{lstlisting}


In the \textit{stats\_path} argument we specify the paths of the metadata for the runs we wish to compare  and in the \textit{stats\_names} the names we assign for the runs.

\begin{figure}[h!]
\begin{subfigure}{.5\textwidth}
  \centering
  \includegraphics[scale=0.38]{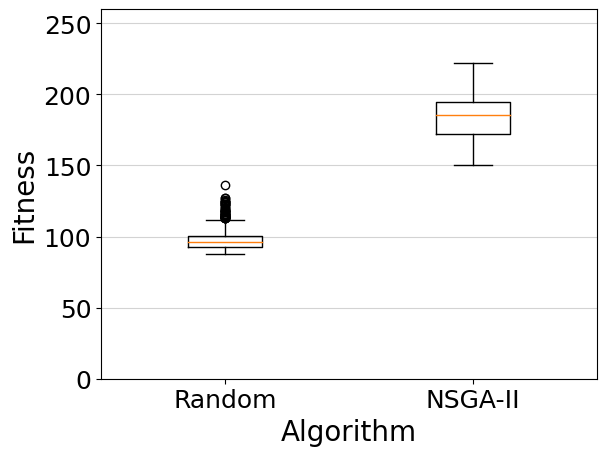}  
  \caption{Scenario fitness}
  \label{fig:boxplot_fit}
\end{subfigure}
\begin{subfigure}{.5\textwidth}
  \centering
  \includegraphics[scale=0.38]{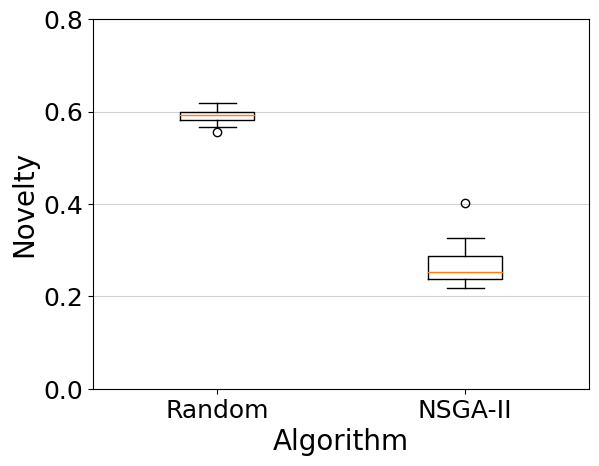}  
  \caption{Scenario diversity}
  \label{fig:boxplot_div}
\end{subfigure}
\caption{Evaluating the NSGA-II algorithm for autonomous robot test case generation}
\label{fig:boxplots}
\end{figure}

\begin{figure}[h!]
\centering
\includegraphics[scale=0.55]{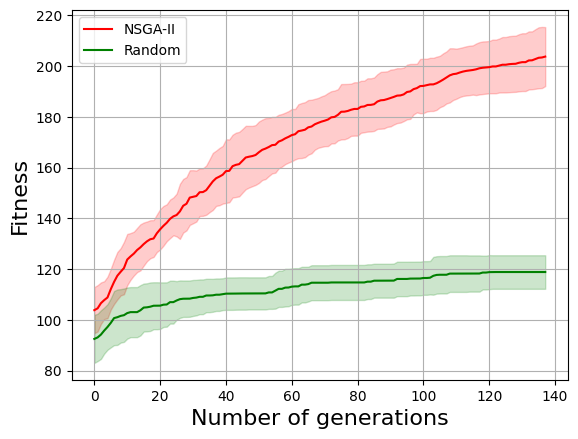}
\caption{Comparing the convergence of NSGA-II and random search for autonomous robot case study}
\label{fig:convergence}
\end{figure}

In Fig.\ref{fig:boxplots} and Fig. \ref{fig:convergence} we can see examples of the outputs produced by the \textit{compare.py} script. Fig. \ref{fig:boxplot_fit} shows the fitness and Fig. \ref{fig:boxplot_div} the diversity of the scenarios in the test suites produced over the specified number of runs.
Fig. \ref{fig:convergence} shows the best values found by the compared search algorithms over the generations.

\section{Illustrative examples}
In this section, we present the summarized results of several test generation case studies using the AmbieGen tool. The full results can be found in our research paper \cite{HUMENIUK2022106936} and the SBST 2022 competition report \cite{gambi2022sbst}.

\begin{figure}[h!]
\begin{subfigure}{0.5\textwidth}
  \centering
  \includegraphics[scale=0.4]{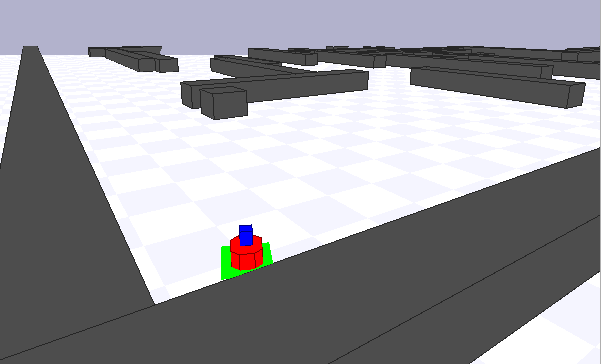}  
  \caption{Executing autonomous robot scenario in the Player/Stage simulator}
  \label{fig:rob1}
\end{subfigure}
\begin{subfigure}{0.5\textwidth}
  \centering
  \includegraphics[scale=0.5]{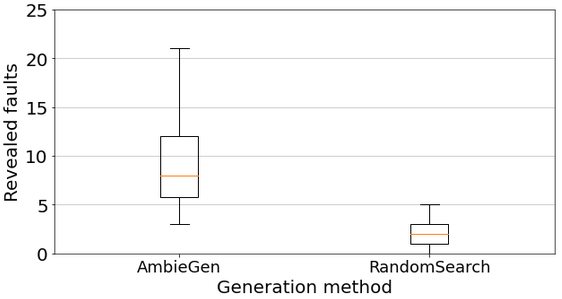}  
  \caption{The number of failures revealed by AmbieGen for the robot case study}
  \label{fig:rob2}
\end{subfigure}
\caption{Using AmbieGen for testing autonomous robot navigation algorithm}
\label{fig:rob}
\end{figure}

We conducted a case study on an autonomous robot with an obstacle avoidance algorithm based on nearness diagrams \cite{minguez2004nearness}. The robot model was a Pioneer 3-AT equipped with a SICK LMS200 laser with a sensing range of 10 meters. The simulations were run in the Player/Stage simulator \cite{kranz2006player}. You can see an illustration of the simulation environment in Fig. \ref{fig:rob1}. We used AmbieGen to generate diverse maps with obstacles to test the robot's performance. We identified several scenarios in which the robot became stuck and failed to reach its goal location. An example of such a scenario can be found in the following video: \href{https://figshare.com/s/7208f6d5ce19e1476474}{Video}.

To evaluate the effectiveness of our tool, we allocated a two-hour budget for AmbieGen to generate test scenarios. The generated scenarios were then passed to the simulator and executed. We repeated the experiment 30 times, using both the NSGA-II and random search configurations of AmbieGen. The average number of failures detected is shown in Fig. \ref{fig:rob2}. On average, AmbieGen detected 9 failures in two hours, compared to 2 failures for random search

\begin{figure}[h!]
\begin{subfigure}{0.5\textwidth}
  \centering
  \includegraphics[scale=0.5]{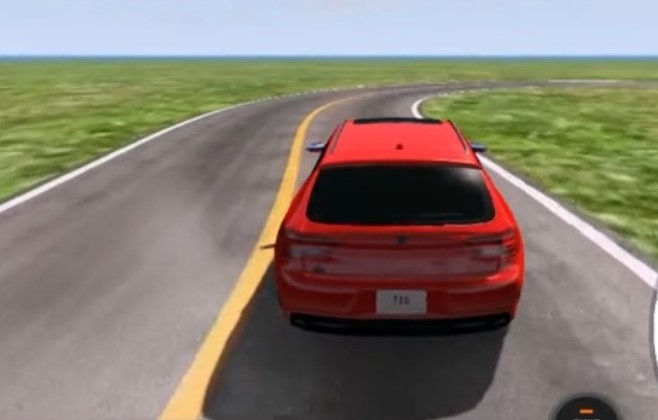}  
  \caption{Executing the LKAS scenario in the BeamNg simulator}
  \label{fig:veh1}
\end{subfigure}
\begin{subfigure}{0.5\textwidth}
  \centering
  \includegraphics[scale=0.8]{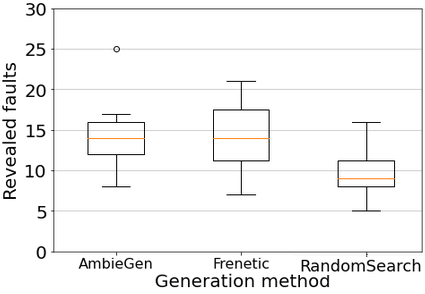}  
  \caption{The number of failures revealed by AmbieGen for the LKAS case study}
  \label{fig:veh2}
\end{subfigure}
\caption{Using AmbieGen to test autonomous vehicle LKAS model}
\label{fig:sim}
\end{figure}

In the second case study, we evaluated the performance of our test generation tool on an autonomous vehicle lane keeping assist system (LKAS) using the BeamNg simulator \cite{beamng}. We used the AmbieGen tool to generate diverse, fault-revealing road topologies, which were then simulated in the BeamNg environment (shown in Fig. \ref{fig:veh1}). During the simulations, we identified a number of scenarios in which the vehicle left its lane (an example of which can be seen in the video at \href{https://share.com/s/b4a096f0a66e0abbe7b1}{Video}).

We ran our tool for a time budget of 2 hours, using the SBST22 competition code pipeline. The failure criterion for the LKAS system was defined as more than 85\% of the car's area leaving the lane. The driving agent had a maximum speed of 70 Km/h. We compared the results of AmbieGen's NSGA-II configuration, Random Search configuration, and the Frenetic tool \cite{castellano2021frenetic}, which was also given a 2-hour time budget for test generation.

As shown in Fig. \ref{fig:veh2}, out of 30 runs, AmbieGen and Frenetic on average produced almost the same number of failures (14), while Random Search produced an average of 9 failures.

The obtained results suggest that AmbieGen could effectively identify failures in the autonomous systems under test.
\section{Impact}
Autonomous systems testing is an important area of research, and finding test scenarios that reveal a diverse range of system failures within a limited time and evaluation budget is a significant challenge \cite{driving}. One of the common solutions is to use evolutionary search to guide the sampling towards more challenging scenarios \cite{gambi2019automatically, riccio2020model}. These search based techniques allow to identify potential failures and improve the overall reliability of the system.

AmbieGen is a test generation tool that uses evolutionary search to generate test scenarios for autonomous systems. Its modular design allows for customization of the initial population generation function, fitness evaluation function, search operators (such as crossover and mutation), and the search algorithm itself. Out of the box, AmbieGen supports testing of autonomous robots and vehicle LKAS systems, and additional systems can be added using the provided implementations as examples.

AmbieGen is a valuable resource for research on search-based test case generation for autonomous systems. Its built-in modules enable easy comparison of different search algorithms and their modifications, based on the quality and diversity of the generated solutions, as well as the convergence of the algorithm over time.

AmbieGen can help answer research questions that are not frequently discussed in the literature, such as:

\begin{itemize}
    \item To what extent the diversity preservation technique A helps improve the diversity of the test suite? The importance of the diversity in test case generation is extensively discussed in the work of Klikovits et al. \cite{klikovits2022does}.
    \item  To what extent does the search operator A helps improve the convergence over the operator B? To what extent the algorithm A outperforms the algorithm B for the test case generation? Improvements to the baseline genetic algorithms implementations can lead to better results, as discussed by Abdessalem et al. \cite{abdessalem2018testing}, where multi-objective population-
based search algorithms and decision tree classification were combined.
    \item What fitness criteria are more relevant for guiding the system towards fault revealing scenarios? This question includes the comparison of the single, multi-objective based search as well surrogate model assisted search.
\end{itemize}
AmbieGen can also be useful in the pursuit of actively studied research questions, where the fault revealing test case generation is required, such as: transferability of failures from simulation to the real world \cite{stocco2022mind}, autonomous system failure prediction \cite{stocco2021confidence}, test case prioritization \cite{arrieta2022some} and others.

AmbieGen has proven its effectiveness in fault revealing by winning this year's edition of the SBST 2022 cyber-physical testing tool competition. Our submission is described in the following article \cite{humeniuk2022ambiegen} and is available at the following link \url{https://github.com/dgumenyuk/tool-competition-av}. We have always kept our tool open sourced and we expect more people to start using it. We welcome all the contributions for expanding our framework.

\section{Conclusions}
In this paper, we present the AmbieGen framework for search based test case generation for autonomous systems, in its public version 0.1.0. We
briefly outline the motivation for developing this framework, its workflow and main functionalities. We also provide illustrative examples for using the tool for autonomous vehicle lane keeping assist system testing and autonomous robot obstacle avoiding algorithm testing. The main features of our tool include:
\begin{itemize}
    \item modular architecture, which allows researchers to easily modify the existing modules, such as initial population generation, crossover, mutation, fitness function as well as introduce  new problems and run experiments;
    \item we provide implementations of test case generation for two systems under test: autonomous vehicle LKAS system and autonomous robot; this implementation includes three search algorithms: random search, single objective genetic algorithm and a two-objective  NSGA-II genetic algorithm; 
    \item our framework is built to be compatible with Pymoo framework \cite{pymoo}, allowing to fully benefit from the Pymoo framework features, such as high number of implemented algorithms in Pymoo.
\end{itemize}

\section{Future Plans}
Our framework currently includes the implementation of two test case generation problems, as well as three algorithms (random search, GA, NSGA-II) for generating test cases. The fitness function is calculated based on a simplified model of the system, and test scenarios are represented as 2D arrays, with each column describing a discrete aspect of the scenario.
In the future, we plan to expand the capabilities of our framework to include: 
\begin{itemize}
    \item new algorithms, especially the ones based on the quality-diversity search \cite{pugh2016quality}
    \item new test case generation problems, for instance more complex test scenarios that include moving pedestrians, other vehicles and traffic signs; 
    \item new fitness functions e.g based on surrogate models of the system under test, as in the work of Ramakrishna et al. \cite{ramakrishna2022risk}, functions based on neuron coverage \cite{pei2017deepxplore} and surprise adequacy \cite{kim2019guiding} dedicated to testing systems  containing neural networks;
    \item add new problem representations, supporting popular scenario specification languages such as SCENIC \cite{fremont2022scenic};
    \item add an integration with popular simulators, for instance CARLA \cite{dosovitskiy2017carla} or LGSVL \cite{rong2020lgsvl}. This will allow to directly evaluate the system model with the generated scenarios. Also the feedback from the simulators could be incorporated in fitness functions for guiding the test scenario sampling.
\end{itemize}

\section*{Acknowledgements}
This work is partly funded by the by the Fonds de Recherche du Québec (FRQ), the Natural Sciences and Engineering Research Council of Canada (NSERC), and the Canadian Institute for Advanced Research (CIFAR).

\bibliographystyle{elsarticle-num}
\bibliography{references}


%
%
%

\end{document}